%% file: main.tex
\documentclass[sigconf]{acmart}
\usepackage{array}
\usepackage{booktabs}
\usepackage{xcolor,colortbl}
\usepackage{makecell}
\AtBeginDocument{%
  }

\acmISBN{978-1-4503-XXXX-X/2018/06}

\begin{document}
\settopmatter{printfolios=false}
\settopmatter{printacmref=false}
\renewcommand\footnotetextcopyrightpermission[1]{}
\pagestyle{empty}
\title{AgentOS: From Application Silos to a Natural Language-Driven Data Ecosystem}


\author{Rui Liu}
\affiliation{
  \institution{University of Kansas}
  \city{Lawrence}
  \country{USA}
}
\email{rayliu@ku.edu}

\author{Tao Zhe}
\affiliation{
  \institution{University of Kansas}
  \city{Lawrence}
  \country{USA}
}
\email{taozhe@ku.edu}

\author{Dongjie Wang}
\authornote{Corresponding author}
\affiliation{
  \institution{University of Kansas}
  \city{Lawrence}
  \country{USA}
}
\email{wangdongjie@ku.edu}

\author{Zijun Yao}
\affiliation{
  \institution{University of Kansas}
  \city{Lawrence}
  \country{USA}
}
\email{zyao@ku.edu}

\author{Kunpeng Liu}
\affiliation{
  \institution{Clemson University}
  \city{Clemson}
  \country{USA}
}
\email{kunpenl@clemson.edu}

\author{Yanjie Fu}
\affiliation{
  \institution{Arizona State University}
  \city{Tempe}
  \country{USA}
}
\email{yanjie.fu@asu.edu}

\author{Huan Liu}
\affiliation{
  \institution{Arizona State University}
  \city{Tempe}
  \country{USA}
}
\email{huanliu@asu.edu}

\author{Jian Pei}
\affiliation{
  \institution{Duke University}
  \city{Durham}
  \country{USA}
}
\email{j.pei@duke.edu}

% Rui Liu, Tao Zhe, Dongjie Wang(corresponding), Zijun Yao, Kunpeng Liu, Yanjie Fu(asu,corresponding), Huan Liu(asu), Jian Pei(duke)

% \affiliation{%

% \affiliation{%

% \affiliation{%

% \affiliation{%

% \affiliation{%

% \affiliation{%

% \affiliation{%

\input{0_abstract}

\maketitle

\input{1_introduction}

\input{2_vision}

\input{3_mining_user_context}

\input{4_challenge}

\input{5_conclusion}

\bibliographystyle{ACM-Reference-Format}
\bibliography{sample-base}

\appendix

\end{document}

%% file: 0_abstract.tex
\begin{abstract}
The rapid emergence of open-source, locally hosted intelligent agents marks a critical inflection point in human-computer interaction. 
Systems such as OpenClaw demonstrate that Large Language Model (LLM)-based agents can autonomously operate local computing environments, orchestrate workflows, and integrate external tools. 
However, within the current paradigm, these agents remain conventional applications running on legacy operating systems originally designed for Graphical User Interfaces (GUIs) or Command Line Interfaces (CLIs). 
This architectural mismatch leads to fragmented interaction models, poorly structured permission management (often described as "Shadow AI"), and severe context fragmentation.
This paper proposes a new paradigm: a Personal Agent Operating System (AgentOS). 
In AgentOS, traditional GUI desktops are replaced by a Natural User Interface (NUI) centered on a unified natural language or voice portal. 
The system core becomes an Agent Kernel that interprets user intent, decomposes tasks, and coordinates multiple agents, while traditional applications evolve into modular Skills-as-Modules enabling users to compose software through natural language rules.
We argue that realizing AgentOS fundamentally becomes a Knowledge Discovery and Data Mining (KDD) problem. 
The Agent Kernel must operate as a real-time engine for intent mining and knowledge discovery. 
Viewed through this lens, the operating system becomes a continuous data mining pipeline involving sequential pattern mining for workflow automation, recommender systems for skill retrieval, and dynamically evolving personal knowledge graphs. 
These challenges define a new research agenda for the KDD community in building the next generation of intelligent computing systems.
\end{abstract}

%% file: 1_introduction.tex
\vspace{-0.2cm}
\section{Introduction: The OpenClaw Moment and the End of the GUI}
\subsection{The Eruption of Local Autonomous Agents}
In early 2026, the computing landscape experienced a rapid shift driven by locally deployed autonomous agents, exemplified by OpenClaw (formerly Moltbot and Clawdbot). 
As an open-source, locally self-hosted AI assistant, OpenClaw quickly gained attention within the developer community, accumulating over 100,000 GitHub stars within weeks. 
It demonstrated a new paradigm of human–computer interaction: an AI agent capable of directly operating personal computing environments on behalf of users~\cite{yao2023react,schick2023toolformer,wang2023voyager}. 
By integrating messaging gateways (e.g., WhatsApp, Telegram, iMessage, and Slack) with the Model Context Protocol (MCP), OpenClaw functions as a persistent background agent capable of reading and writing files, executing terminal commands, managing calendars, and browsing the web~\cite{OpenClawRepo2026,Bonatti2025ComputerUsingAgents}. 
Developers often describe it as ``Claude with hands,'' highlighting that it is not merely a conversational interface but a continuously running agent interacting directly with the operating system. 
Through an autonomous \textit{agentic loop}, it maintains long-term memory and executes long-horizon tasks with minimal supervision~\cite{Mei2025AIOS}. 
Meanwhile, proprietary systems—such as Anthropic's Claude Code Remote Control and Perplexity Computer—illustrate growing demand for scheduled task execution, deep contextual integration, and remote AI orchestration in enterprise and developer workflows.
However, the rapid adoption of such agents exposes a fundamental limitation of modern computing architectures: they still operate as user-space processes on legacy operating systems (e.g., Windows, macOS, and Linux) designed for human-driven graphical interfaces and isolated application silos rather than continuously operating autonomous agents.

\begin{figure}[!t]
    \centering
    \includegraphics[width=\linewidth]{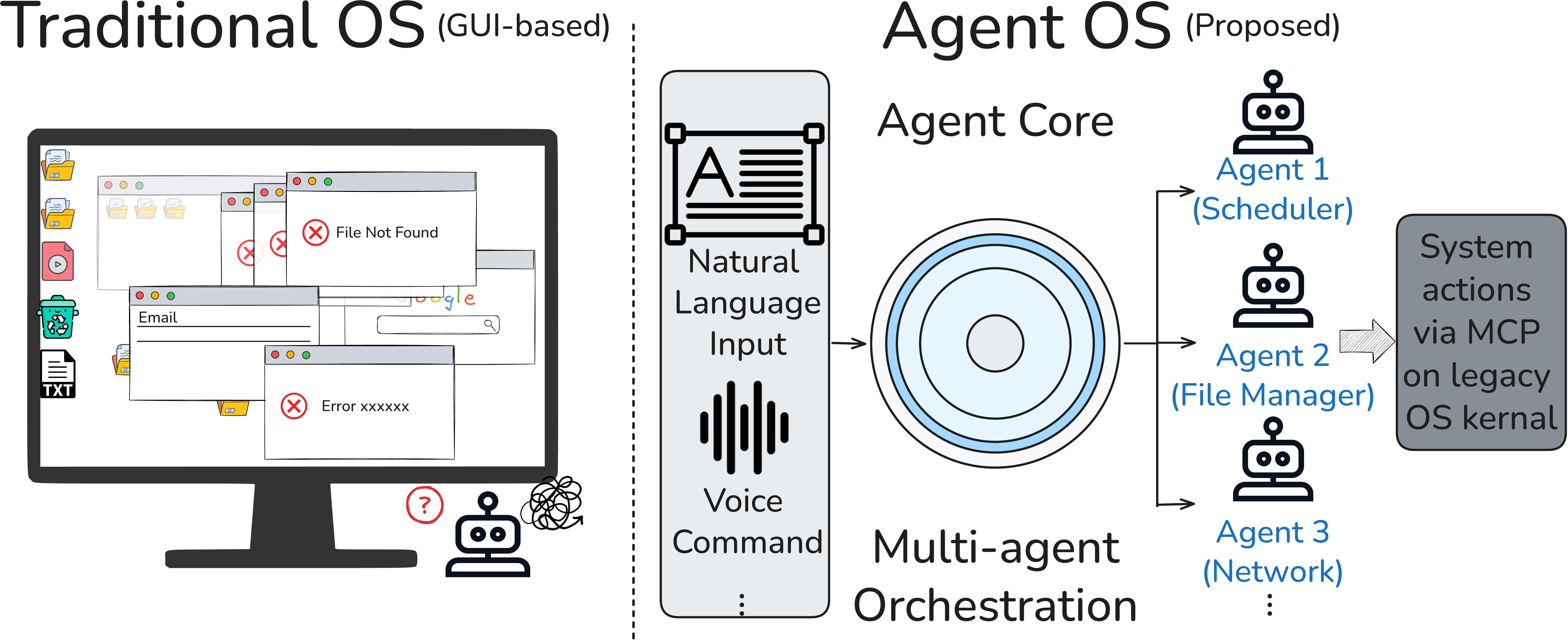}
    \caption{Paradigm Shift from GUI-Based Operating Systems to AgentOS with Multi-Agent Orchestration and Natural Language Interface.}
    \vspace{-0.5cm}
\end{figure}

\vspace{-0.3cm}
\subsection{The "Shadow AI" Crisis and the "Screen-as-Interface" Bottleneck}
Forcing autonomous agents to operate within legacy operating systems creates a phenomenon analogous to ``Shadow IT,'' which we term \textit{Shadow AI}. 
Because these systems lack semantic understanding of agent actions, agents rely on brittle interaction workarounds. 
Consequently, many mobile and desktop agents operate under a ``Screen-as-Interface'' paradigm, using visual scraping (pixels-to-text) or simulated GUI interactions (e.g., mouse clicks and keystrokes) to perform tasks~\cite{nakano2022webgpt}. 
This GUI-centric interaction model is fundamentally ill-suited for autonomous agents for several reasons:

\begin{itemize}

\item \textbf{Loss of Semantic Information:} 
GUIs are designed for human visual interpretation, presenting visual outputs while obscuring underlying structured data and metadata. 
Agents that ``read'' the screen therefore lose semantic context, often leading to reasoning errors~\cite{deng2023mind2webgeneralistagentweb,zheng2024seeact}.

\item \textbf{Fragile Execution Paths:} 
Interface layouts frequently change. When applications update their interfaces, agents relying on visual markers or fixed screen coordinates may fail catastrophically~\cite{zhou2024webarenarealisticwebenvironment}.

\item \textbf{Runaway Security and Permissions:} 
Legacy operating systems manage permissions at the application level (e.g., file system access). 
Once an autonomous agent receives such permissions, the system cannot reliably distinguish legitimate actions from malicious behaviors (e.g., indirect prompt injection leading to data exfiltration)~\cite{Su2025AutonomySecuritySurvey,Chaturvedi2025AIP}.

\end{itemize}

\subsection{Paradigm Shift: From GUI to Natural User Interfaces}
Historical trends in computing indicate a transition from graphical user interfaces (GUIs) toward natural user interfaces (NUIs). 
Traditional operating systems are largely passive, responding to explicit user commands through deterministic program logic. 
Future computing environments, however, require proactive and probabilistic systems capable of inferring user intent from ambiguous and multimodal interactions~\cite{Sun2016IntentTracking}. 
Addressing the growing mismatch between intelligent software and legacy system architectures therefore requires a clean-slate architectural redesign. 
In this paradigm, the traditional operating system kernel is encapsulated beneath a new intelligent layer—the \textit{Agent Kernel}—which becomes the primary interface between users and the system, transforming the computer from isolated applications into a unified natural language–driven computational platform.

%% file: 2_vision.tex
\section{Vision: The Architectural Reconstruction of AgentOS}
AgentOS represents a structural rethinking of the computing environment~\cite{packer2024memgptllmsoperatingsystems}. 
Rather than operating as an application on top of an existing system, AgentOS redefines the operating system itself. 
Under this paradigm, traditional GUI-driven desktop environments and isolated application layers are replaced by an agent-centric architecture. 
The Agent Kernel acts as the primary system interface, translating natural language intent into deterministic actions executed by sandboxed, user-defined skill modules.

\begin{figure}[!t]
    \centering
    \includegraphics[width=0.8\linewidth]{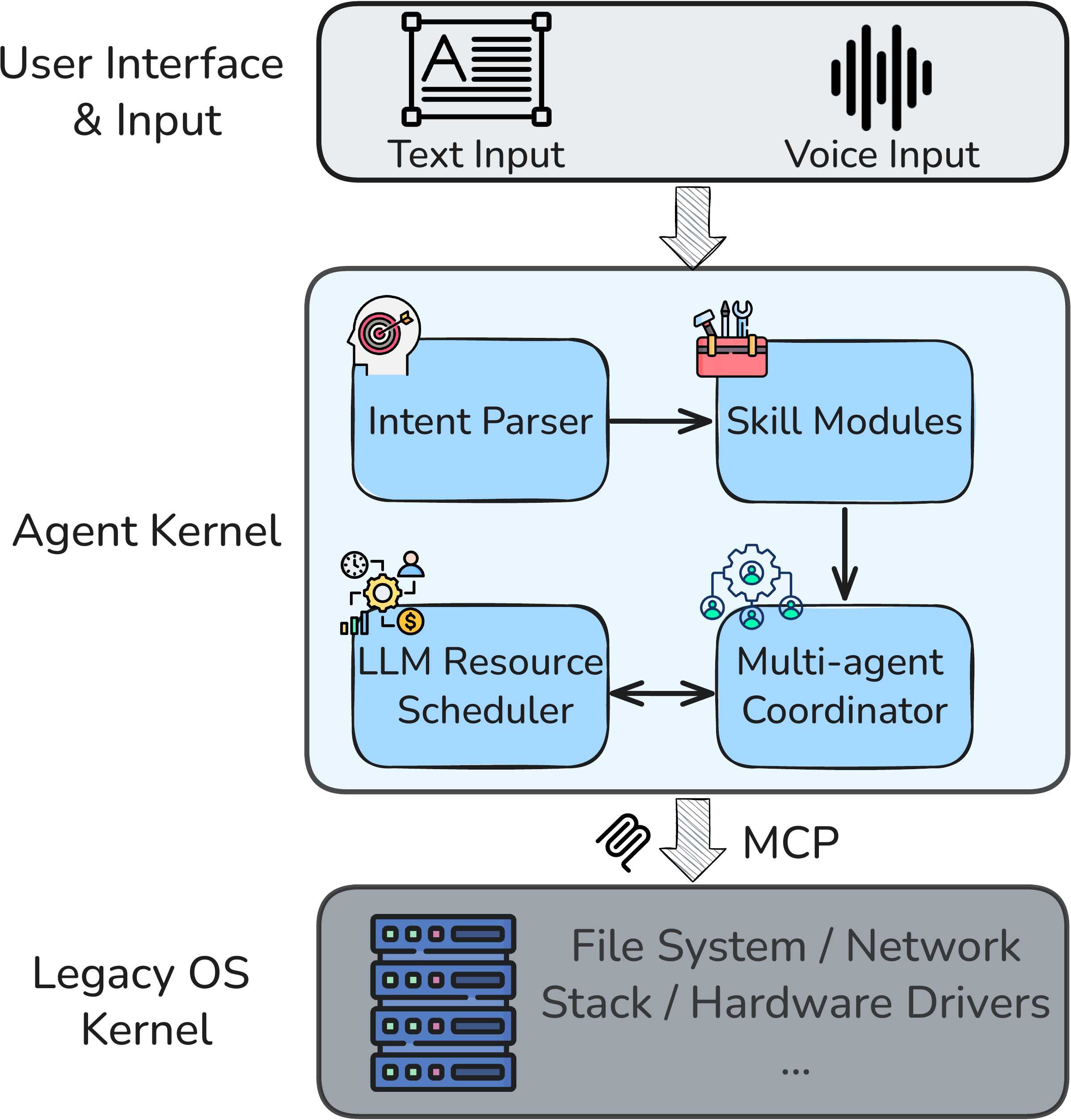}
    \caption{Layered Architecture of AgentOS: Single Port Interface, Agent Kernel, and Legacy Infrastructure Abstraction.}
\end{figure}

\input{figures/LegacyOSvsAgentOS}

\subsection{The Single Port: The Death of the Desktop}
A defining user-facing characteristic of AgentOS is the replacement of the traditional desktop metaphor with a unified interaction gateway referred to as \textit{the Single Port}~\cite{kaushik2014naturaluserinterfacestrend,880078}. 
Conventional interface elements such as icons, menus, taskbars, and window management become secondary interaction mechanisms rather than the primary interface.
In the default state, the system remains in a minimalist standby mode, exposing a persistent multimodal interface that accepts voice, text, and contextual signals. 
Users interact with the system primarily through natural language instructions. 
Visual interfaces are generated only when necessary—for example, when resolving ambiguous instructions or presenting inherently visual outputs such as charts, maps, or videos.
By centralizing interaction into a single semantic interface, AgentOS significantly reduces the cognitive overhead associated with navigating across multiple applications and interface contexts.

\subsection{Agent Kernel Architecture: From Process Scheduling to Intent Orchestration}
At the core of AgentOS lies the Agent Kernel, which abstracts the complexities of physical hardware and legacy operating systems. 
Building upon early AIOS (LLM-based Agent Operating System) frameworks proposed in 2024, the Agent Kernel integrates agent applications with system resources to support coordinated execution, resource management, and security enforcement.
The Agent Kernel exposes two primary interfaces:

\begin{itemize}

\item \textbf{Northbound Interface (Intent Translation):} 
Facing the user, this interface performs continuous semantic parsing, transforming ambiguous and multimodal human inputs (e.g., voice commands, gestures, or contextual queries) into structured, machine-executable intents. 
It maintains persistent user context, manages conversational state, and handles interruptions during long-horizon tasks.

\item \textbf{Southbound Interface (Multi-Agent Orchestration):} 
Facing the underlying computing infrastructure, this interface activates once intent structuring is complete. 
An internal Multi-Agent System (MAS) decomposes user requests into executable sub-tasks and dispatches them through the Model Context Protocol (MCP) to interact with the ``invisible'' legacy OS kernel, including the file system, network stack, and hardware drivers~\cite{Wang2026InternetOfAgents}.

\end{itemize}
Crucially, the Agent Kernel must also perform large-scale LLM resource scheduling. 
Analogous to how traditional kernels multiplex CPU time across processes, the Agent Kernel allocates limited LLM resources—including context windows, token budgets, and API rate limits—across multiple concurrent agent threads. 
This scheduling layer prevents out-of-memory failures and maintains system throughput under high concurrency.

\begin{figure}[!t]
    \vspace{-0.2cm}
    \centering
    \includegraphics[width=\linewidth]{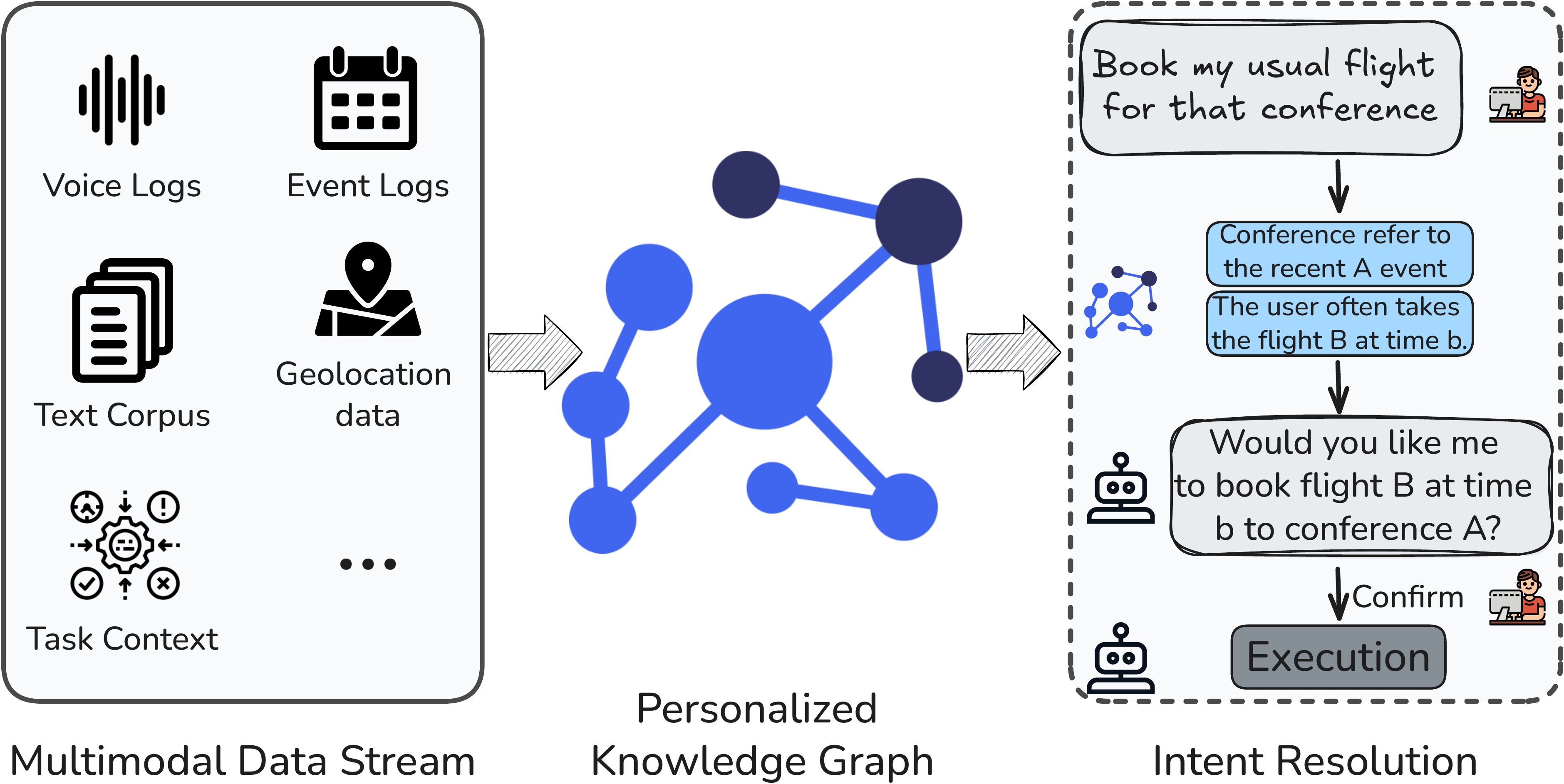}
    \caption{Multimodal Intent Mining and Personal Knowledge Graph Construction within the Agent Kernel.}
    \vspace{-0.6cm}
\end{figure}

\subsection{Skills as User-Defined Software}
Within the AgentOS ecosystem, the traditional notion of pre-packaged applications becomes significantly less central. 
Instead of installing monolithic software packages from centralized app stores, functionality is expressed through modular \textit{Skills-as-Modules}. 
These skills represent reusable automation logic that can be composed dynamically to satisfy user goals.
Users can define such skills directly through natural language specifications. 
For example, a user may specify: 
``Whenever I receive an email from the finance director containing a PDF invoice, extract the total amount, verify the corresponding item in my budget spreadsheet, and if correct, draft a payment authorization.''
The Agent Kernel converts this natural language rule into a structured skill module and persistently stores it within the user environment. 
Each skill operates as a composable microservice that can be invoked independently or combined with other skills to form more complex workflows.
This paradigm resembles emerging agentic AI architectures in which specialized agents divide responsibilities and collaboratively execute complex tasks~\cite{Liu2026AgenticUrbanPlanning,liu2026cityeditinghierarchicalagentic,zhe2025constraintawarerouterecommendationnatural}. 
Over time, the operating system evolves into a highly personalized ecosystem of skills, yielding a computing experience tailored to the individual user's workflows and preferences.

%% file: figures/LegacyOSvsAgentOS.tex
\begin{table*}[t]
\centering
\renewcommand{\arraystretch}{1.2} 
\setlength{\tabcolsep}{8pt} 

% 定义与浅绿色对应的同色系浅蓝色
\definecolor{RowTintBlue}{RGB}{240,248,255}

\begin{tabular}{
  >{\raggedright\arraybackslash}m{3.6cm}
  >{\raggedright\arraybackslash}m{5.3cm}
  >{\raggedright\arraybackslash}m{5.3cm}
}
\toprule
\textbf{Architecture Layer} & \textbf{Legacy OS} & \textbf{AgentOS} \\
\midrule

\rowcolor{RowTintBlue}
\textbf{Human--Computer Interface (HCI)}
& GUI / Desktop Window Manager
& Single Natural Language Port (Voice / Text) \\
\addlinespace[4pt]

\textbf{Application Layer}
& Isolated Third-Party Applications (Apps)
& User-Defined Skill Modules (Skills) \\
\addlinespace[2pt]

\rowcolor{RowTintBlue}
\textbf{Core Management Engine}
& Process Scheduling, Memory Management
& Intent Parser, Multi-Agent Coordinator, LLM Resource Scheduler \\
\addlinespace[2pt]

\textbf{Underlying Interaction}
& System Calls (Syscalls), POSIX
& Model Context Protocol (MCP), Semantic API \\
\addlinespace[2pt]

\rowcolor{RowTintBlue}
\textbf{Hardware Abstraction}
& Physical Hardware Drivers
& Legacy OS Kernel (Invisible Infrastructure) \\
\bottomrule
\end{tabular}

\vspace{0.2cm}
\caption{Comparison between Legacy OS and AgentOS architectures.}
\label{tab:agentos_vs_legacy}
\vspace{-0.3cm}
\end{table*}

%% file: 3_mining_user_context.tex
\section{Mining User Context: Why the Future OS is a Data Mining Problem}
While the architectural vision of AgentOS is disruptive, its technical realization depends fundamentally on advances in Knowledge Discovery and Data Mining (KDD). 
Traditional operating systems are deterministic engineering systems built on static logic, rigid file hierarchies, and explicit machine instructions. 
AgentOS, in contrast, operates in a probabilistic environment: it must infer user intent, resolve linguistic ambiguity, and adapt to evolving behavioral patterns. 
Consequently, the core challenge of AgentOS shifts from conventional systems engineering to real-time data mining, where the operating system continuously extracts structured intent from large-scale unstructured interaction data.

\input{figures/Evaluationframework}

% \vspace{-0.2cm}
\subsection{Intent Mining \& Context Awareness}

Traditional operating systems are reactive, responding only to explicit user commands. 
AgentOS, however, must proactively infer user intent from ambiguous and incomplete inputs. 
Without contextual grounding, requests such as ``Book my usual flight for that conference'' remain ambiguous for machines~\cite{Sun2016IntentTracking}.

\begin{itemize}

\item \textbf{KDD Challenge: Dynamic Personal Knowledge Graphs}  
To resolve such ambiguity, AgentOS requires continuous construction and querying of a Personal Knowledge Graph (PKG)~\cite{Skjaeveland2024PKGSurvey}. 
The PKG serves as a semantically rich, user-controlled knowledge base capturing personal data, preferences, social relationships, and behavioral history to support personalized reasoning.

\item As a continuous intent mining engine, AgentOS processes multimodal interaction streams—including voice logs, screen context, and geolocation signals—and applies Natural Language Processing (NLP) and relational extraction techniques to update the PKG in real time. 
When handling ambiguous instructions, the Agent Kernel performs graph-augmented reasoning over the PKG to infer implicit user preferences from historical behavior~\cite{Kim2026Persona2Web}.

\item The \textit{Zero-shot-to-head-shot hyperpersonalization} (Z2H2) framework offers one implementation pathway~\cite{Dandeniya2025Hyperpersonalization}. 
It categorizes user data into explicit, implicit, passive, and derivative modalities, enabling AgentOS to evolve from cold-start prompts toward personalized interaction through Retrieval-Augmented Generation and behavioral profiling. 
Without robust intent mining and PKG integration, agents risk repeatedly interrupting users for clarification or executing incorrect actions.

\end{itemize}

\vspace{0.4cm}
\subsection{Skill Retrieval \& Recommendation}
As traditional App Stores evolve into large-scale Skill Ecosystems, users will accumulate extensive collections of highly granular and overlapping skill modules. 
The operating system must therefore identify which specific skill—or sequence of skills—best matches a given task. 
This process resembles tool orchestration in agentic AI systems, where complex workflows are constructed by dynamically selecting and composing specialized tools or modules~\cite{zhe2026robustefficienttoolorchestration}.

\vspace{0.5cm}
\begin{itemize}

    \item \textbf{KDD Challenge: Hyperscale Recommender Systems for Software and Logic}  
    This problem can be framed as a large-scale recommender system task~\cite{Zhao2021JobMatching}. 
    Unlike traditional recommendation settings (e.g., movies or e-commerce), AgentOS must recommend executable logic—such as skill modules or code sequences—based on complex and multidimensional user context.
    
    \item A natural solution is a \textit{Two-Tower Recommendation Architecture}. 
    The ``User Tower'' encodes natural language queries together with real-time PKG context into a high-dimensional \textit{context embedding}~\cite{Wang2025IntentLLMRec}. 
    The ``Skill Tower'' represents skill modules by embedding functional metadata, code semantics, and execution constraints into a shared representation space. 
    Similarity search (e.g., cosine similarity) is then used to retrieve the most relevant skill modules~\cite{Tang2025ExplainableJobRec}.
    
    \item Recommendation quality can be further improved through Reinforcement Learning (RL) and collaborative filtering. 
    User feedback—such as task abortion or manual correction—provides negative signals that allow the system to update skill embeddings and improve future retrieval accuracy~\cite{zou2019feedrec,zhao2018pagewise,Lin_2024}.

\end{itemize}

% \vspace{-0.3cm}
\subsection{Action Sequence Mining and Optimization}
An AgentOS operating through a multi-agent system will generate large volumes of interaction data. 
As agents interact with file systems, network APIs, and web environments, they produce continuous \textit{action traces}—sequences of system calls, API requests, and navigation steps executed to complete tasks~\cite{Leno2022DataTransferRoutines}.

\begin{itemize}
    \item \textbf{KDD Challenge: Sequential Pattern Mining (SPM)}  
    To improve efficiency over time, the operating system must apply Sequential Pattern Mining (SPM) techniques to analyze large-scale UI and system logs. 
    SPM algorithms are well suited for discovering frequently occurring subsequences in temporal interaction data.
    
    \item By mining action traces, the system can identify repetitive and high-latency workflows. 
    When frequent patterns are detected, the Agent Kernel can automatically synthesize optimized macros or background services that replace multi-step interaction sequences with direct execution.
    
    \item A key challenge lies in handling the noise inherent in real-world computing environments, including fluctuating action spaces and unpredictable external responses. 
    Robust sequence mining therefore requires integrating action models and filtering heuristics to distinguish meaningful behavioral patterns from spurious interaction traces.
\end{itemize}

\subsection{Evaluation \& Benchmarks}
Transitioning to a probabilistic and highly personalized operating system challenges traditional software evaluation metrics. 
Performance can no longer be assessed solely through CPU utilization, memory efficiency, or binary unit tests. 
Because system behavior varies across users and contexts, the notion of ``correctness'' becomes inherently subjective and context-dependent.

\begin{itemize}

    \item \textbf{KDD Challenge: Quantifying Intent Alignment}  
    Evaluating AgentOS requires new benchmarks centered on user satisfaction and \textit{Intent Alignment} (IA)~\cite{Guan2026MultiTurnSurvey,Shukla2025AgenticEvaluationSurvey}. 
    IA measures the semantic gap between a user's latent goal and the actions executed by the agent.
    
    \item Recent research has begun addressing this challenge through new evaluation frameworks. 
    For example, the Tri-Agent framework introduces a clarification agent, a response agent, and an evaluator agent (LLM-as-a-judge) to assess conversational efficiency, disambiguation, and intent alignment~\cite{Zhao2025TriAgent}. 
    Similarly, initiatives such as the Agentic Benchmark Checklist (ABC) enforce structured multi-turn task settings to prevent overestimation of agent capabilities~\cite{Zhu2025AgenticBenchmarks}.
    
    \item Future evaluation frameworks should integrate simulation environments (e.g., AndroidArena) together with learning-based reward modeling methods such as Markovian Intrinsic Reward Adjustment (MIRA) to measure cognitive agency, tool proficiency, and long-horizon task performance~\cite{10.1145/3711896.3736570}.

\end{itemize}

%% file: figures/Evaluationframework.tex
\begin{table*}[!ht]
\centering
\renewcommand{\arraystretch}{1.2} 
\setlength{\tabcolsep}{8pt} 

\definecolor{RowTint}{RGB}{240,252,248}

\begin{tabular}{
  >{\raggedright\arraybackslash}m{3.2cm}
  >{\raggedright\arraybackslash}m{5.2cm}
  >{\raggedright\arraybackslash}m{5.2cm}
}
\toprule
\textbf{Evaluation Dimension} & \textbf{Legacy OS Metrics} & \textbf{AgentOS Metrics} \\
\midrule

\rowcolor{RowTint}
\textbf{Primary Objective}
& System stability and resource utilization
& User intent fulfillment and complex workflow automation \\
\addlinespace[4pt]

\textbf{Key Performance Indicators (KPIs)}
& CPU load, memory faults, disk I/O
& Intent Alignment (IA), task completion rate, tool invocation accuracy \\
\addlinespace[2pt]

\rowcolor{RowTint}
\textbf{Fault Tolerance \& Measurement}
& Crash logs, kernel panics, exception traces
& Hallucination rate, context drift, disambiguation failure rate \\
\addlinespace[2pt]

\textbf{Benchmarking Methodology}
& Unit testing, integration testing, stress testing
& Tri-Agent evaluation, AndroidArena simulations, Agentic Benchmark Checklist (ABC) \\
\addlinespace[2pt]

\rowcolor{RowTint}
\textbf{System Evolution Mechanism}
& Static; manual developer patches and upgrades
& Dynamic; self-evolving via Sequential Pattern Mining (SPM) and MIRA \\

\bottomrule
\end{tabular}
\vspace{0.2cm}
\caption{Evaluation framework comparison between Legacy OS and AgentOS.}
\label{tab:evaluation_comparison}
% \vspace{-0.3cm}
\end{table*}

%% file: 4_challenge.tex
% \vspace{-0.1cm}
\section{Challenges and Risks: Governing a Probabilistic OS}
Delegating core system control to autonomous probabilistic agents introduces significant systemic risks. 
Because AgentOS operates under inherent uncertainty, robust security architectures and fault-tolerance mechanisms become essential.

% \vspace{-0.15cm}
\subsection{Privacy, Security, and the Semantic Firewall}
Legacy operating systems rely on static permissions and deterministic Access Control Lists (ACLs), where applications either possess access to a resource or they do not. 
In AgentOS, however, agents require broad system-level access to coordinate tasks across multiple resources. 
Security boundaries must therefore shift from verifying \textit{who} requests data to evaluating the \textit{semantic intent} of the request.
Recent studies reveal structural vulnerabilities in system-integrated AI agents. 
For example, an attacker may embed an Indirect Prompt Injection in an email, instructing the agent to retrieve sensitive data such as SSH keys~\cite{Zhang2025SemanticInjection} . 
An agent with unrestricted file-system access may execute such instructions without verification, highlighting the susceptibility of multimodal agents to jailbreak attacks, prompt injections, and adversarial inputs.
To mitigate these risks, AgentOS requires a \textit{Semantic Firewall}~\cite{CastroMaldonado2026SemanticFirewall}. 
Integrated within the Agent Kernel, it functions as a real-time text and data mining security layer that monitors information flows into and out of the LLM core. 
External inputs must undergo sanitization and intent analysis before execution; if malicious intent is detected, the firewall blocks the action and alerts the user.
Core capabilities include:

\begin{itemize}

\item \textbf{Input Sanitization and Intent Vetting:} 
Text mining techniques detect adversarial prompts and jailbreak attempts in incoming data streams or RAG-retrieved documents.

\item \textbf{Taint-Aware Memory and Cognitive Integrity:} 
Inspired by the experimental ``Aura'' OS architecture~\cite{Zou2026BlindGods}, the firewall labels data from untrusted sources as \textit{tainted} and prevents it from triggering high-privilege operations (e.g., password changes or financial transactions).

\item \textbf{Real-Time Data Loss Prevention (DLP):} 
Outbound agent actions are analyzed to detect and prevent the leakage of sensitive entities (e.g., SSNs, API keys, or financial records).

\end{itemize}

% \vspace{-0.5cm}
\subsection{Hallucination Control and System Fault Tolerance}
Beyond malicious attacks, the probabilistic nature of LLMs means the OS is inevitably prone to hallucinations and reasoning errors~\cite{Huang_2025,10.1145/3571730}. 
A traditional OS rarely deletes a critical directory without explicit, hard-coded instructions. 
Conversely, an AgentOS misunderstanding a vague command like "clean up my workspace" could irreversibly delete project files or corrupt system configurations.
To ensure fault tolerance, the Agent Kernel must operate within strict sandboxes, utilizing advanced virtualization and containerization to isolate high-risk operations. 
Most critically, the OS requires a robust, system-level "State Rollback" mechanism. 
By leveraging underlying file systems (like ZFS or Btrfs variants), the OS must maintain fine-grained snapshots. 
If an action trajectory is deemed erroneous (either automatically detected via meta-reflection algorithms or reported by the user), the Agent Kernel must reversely traverse the sequence, undoing state changes to restore absolute system integrity in milliseconds.

%% file: 5_conclusion.tex
% \vspace{-0.2cm}
\section{Conclusion}
The rapid emergence of local autonomous agents, exemplified by systems such as OpenClaw, signals a fundamental shift in personal computing. 
However, deploying these probabilistic systems on legacy operating systems designed for graphical interfaces is increasingly inadequate, leading to fragmented contextual reasoning, new security risks, and limited autonomy.
AgentOS rethinks the operating system paradigm. 
By introducing a unified natural language interface (the Single Port), an Agent Kernel for intent orchestration, and dynamically composable Skill Modules, it transforms the computer from isolated applications into a coherent intent-driven system.
Realizing this vision is not merely a systems engineering challenge but fundamentally a problem of Knowledge Discovery and Data Mining (KDD). 
The effectiveness of the Agent Kernel depends on algorithmic advances that enable machines to infer and model user intent from large-scale interaction data.
Key enabling technologies include Personal Knowledge Graphs (PKGs) for contextual reasoning, Sequential Pattern Mining (SPM) for workflow discovery, recommender systems for skill retrieval, and semantic security mechanisms such as the Semantic Firewall. 
Together, these components redefine the operating system as a continuous data mining pipeline that converts unstructured interactions into executable intent.
Ultimately, the operating system of the future will be defined by its ability to interpret and operationalize human intent, evolving into a continuously learning data mining system.

%% file: sample-base.bib
@String{Computing = "Computing" }

@String{Computer = "{IEEE} Computer" }

@article{Zou2026BlindGods,
  author  = {Zhenhua Zou and Sheng Guo and Qiuyang Zhan and Lepeng Zhao and Shuo Li and Qi Li and Ke Xu and Mingwei Xu and Zhuotao Liu},
  title   = {Blind Gods and Broken Screens: Architecting a Secure, Intent-Centric Mobile Agent Operating System},
  year    = {2026},
  journal = {arXiv preprint arXiv:2602.10915},
  url     = {https://arxiv.org/abs/2602.10915}
}

@article{Mei2025AIOS,
  author  = {Kai Mei and Xi Zhu and Wujiang Xu and Wenyue Hua and Mingyu Jin and Zelong Li and Shuyuan Xu and Ruosong Ye and Yingqiang Ge and Yongfeng Zhang},
  title   = {AIOS: LLM Agent Operating System},
  year    = {2025},
  journal = {arXiv preprint arXiv:2403.16971},
  url     = {https://arxiv.org/abs/2403.16971}
}

@article{Bonatti2025ComputerUsingAgents,
  author  = {Piero A. Bonatti and John Domingue and Anna Lisa Gentile and Andreas Harth and Olaf Hartig and Aidan Hogan and Katja Hose and Ernesto Jimenez-Ruiz and Deborah L. McGuinness and Chang Sun and Ruben Verborgh and Jesse Wright},
  title   = {Towards Computer-Using Personal Agents},
  year    = {2025},
  journal = {arXiv preprint arXiv:2503.15515},
  url     = {https://arxiv.org/abs/2503.15515}
}

@article{Wang2026InternetOfAgents,
  author  = {Yuntao Wang and Shaolong Guo and Yanghe Pan and Zhou Su and Fahao Chen and Tom H. Luan and Peng Li and Jiawen Kang and Dusit Niyato},
  title   = {Internet of Agents: Fundamentals, Applications, and Challenges},
  journal = {IEEE Transactions on Cognitive Communications and Networking},
  year    = {2026},
  volume  = {12},
  pages   = {4476--4501},
  doi     = {10.1109/TCCN.2025.3623369}
}

@article{Skjaeveland2024PKGSurvey,
  author  = {Martin G. Skj{\ae}veland and Krisztian Balog and Nolwenn Bernard and Weronika {\L}ajewska and Trond Linjordet},
  title   = {An Ecosystem for Personal Knowledge Graphs: A Survey and Research Roadmap},
  journal = {AI Open},
  year    = {2024},
  volume  = {5},
  pages   = {55--69},
  doi     = {10.1016/j.aiopen.2024.01.003}
}

@inproceedings{Sun2016IntentTracking,
  author    = {Yu Sun and Nicholas Jing Yuan and Yingzi Wang and Xing Xie and Kieran McDonald and Rui Zhang},
  title     = {Contextual Intent Tracking for Personal Assistants},
  booktitle = {Proceedings of the 22nd ACM SIGKDD International Conference on Knowledge Discovery and Data Mining},
  year      = {2016},
  pages     = {273--282},
  doi       = {10.1145/2939672.2939676}
}

@article{Kim2026Persona2Web,
  author  = {Serin Kim and Sangam Lee and Dongha Lee},
  title   = {Persona2Web: Benchmarking Personalized Web Agents for Contextual Reasoning with User History},
  year    = {2026},
  journal = {arXiv preprint arXiv:2602.17003},
  url     = {https://arxiv.org/abs/2602.17003}
}

@article{Dandeniya2025Hyperpersonalization,
  author  = {Kanishka Dandeniya and Sam Saltis and Shalinka Jayatilleke and Nishan Mills and Harsha Moraliyage and Daswin De Silva and Milos Manic},
  title   = {Zero-Shot to Head-Shot: Hyperpersonalization in the Age of Generative AI},
  journal = {Applied System Innovation},
  year    = {2025},
  volume  = {8},
  number  = {6},
  pages   = {186},
  doi     = {10.3390/asi8060186}
}

@article{Zhao2021JobMatching,
  author  = {Jing Zhao and Jingya Wang and Madhav Sigdel and Bopeng Zhang and Phuong Hoang and Mengshu Liu and Mohammed Korayem},
  title   = {Embedding-Based Recommender System for Job to Candidate Matching on Scale},
  year    = {2021},
  journal = {arXiv preprint arXiv:2107.00221},
  url     = {https://arxiv.org/abs/2107.00221}
}

@article{Tang2025ExplainableJobRec,
  author  = {Fang Tang and Renqi Zhu and Feng Yao and Junzhi Wang and Lailong Luo and Bo Li},
  title   = {Explainable Person--Job Recommendations: Challenges, Approaches, and Comparative Analysis},
  journal = {Frontiers in Artificial Intelligence},
  year    = {2025},
  doi     = {10.3389/frai.2025.1660548}
}

@inproceedings{Wang2025IntentLLMRec,
  author    = {Yu Wang and Lei Sang and Yi Zhang and Yiwen Zhang},
  title     = {Intent Representation Learning with Large Language Model for Recommendation},
  booktitle = {Proceedings of the 48th International ACM SIGIR Conference on Research and Development in Information Retrieval},
  year      = {2025},
  pages     = {1870--1879},
  doi       = {10.1145/3726302.3730011}
}

@article{Leno2022DataTransferRoutines,
  author  = {Volodymyr Leno and Adriano Augusto and Marlon Dumas and Marcello La Rosa and Fabrizio Maria Maggi and Artem Polyvyanyy},
  title   = {Discovering Data Transfer Routines from User Interaction Logs},
  journal = {Information Systems},
  year    = {2022},
  volume  = {107},
  pages   = {101916},
  doi     = {10.1016/j.is.2021.101916}
}

@article{Zhu2025AgenticBenchmarks,
  author  = {Yuxuan Zhu and Tengjun Jin and Yada Pruksachatkun and et al.},
  title   = {Establishing Best Practices for Building Rigorous Agentic Benchmarks},
  year    = {2025},
  journal = {arXiv preprint arXiv:2507.02825},
  url     = {https://arxiv.org/abs/2507.02825}
}

@article{Zhao2025TriAgent,
  author  = {Yikai Zhao},
  title   = {A Tri-Agent Framework for Evaluating and Aligning Question Clarification Capabilities of Large Language Models},
  year    = {2025},
  url     = {https://www.amazon.science/publications/a-tri-agent-framework-for-evaluating-and-aligning-question-clarification-capabilities-of-large-language-models}
}

@article{Guan2026MultiTurnSurvey,
  author  = {Shengyue Guan and Jindong Wang and Jiang Bian and Bin Zhu and Jian-guang Lou and Haoyi Xiong},
  title   = {Evaluating LLM-Based Agents for Multi-Turn Conversations: A Survey},
  year    = {2026},
  journal = {arXiv preprint arXiv:2503.22458},
  url     = {https://arxiv.org/abs/2503.22458}
}

@article{Shukla2025AgenticEvaluationSurvey,
  author  = {Manish Shukla},
  title   = {Evaluation and Benchmarking of Generative and Agentic AI Systems: A Comprehensive Survey},
  journal = {Preprints},
  year    = {2025},
  doi     = {10.20944/preprints202512.1421.v1}
}

@inproceedings{Chaturvedi2025AIP,
  author    = {Saket Sanjeev Chaturvedi and Gaurav Bagwe and Lan Emily Zhang and Xiaoyong Yuan},
  title     = {AIP: Subverting Retrieval-Augmented Generation via Adversarial Instructional Prompt},
  booktitle = {Proceedings of EMNLP 2025},
  year      = {2025},
  pages     = {15861--15878},
  doi       = {10.18653/v1/2025.emnlp-main.801}
}

@article{Zhang2025SemanticInjection,
  author  = {Yi Zhang and Jantan Aman},
  title   = {Targeted Injection Attack Toward the Semantic Layer of Large Language Models},
  journal = {Frontiers in Computer Science},
  year    = {2025},
  doi     = {10.3389/fcomp.2025.1683495}
}

@article{CastroMaldonado2026SemanticFirewall,
  author  = {Victor Castro-Maldonado and Marco A. Aceves-Fernandez and Luis R. Garcia-Noguez and Jesus C. Pedraza-Ortega},
  title   = {Semantic Firewalls with Online Ensemble Learning for Secure Agentic RAG Systems in Financial Chatbots},
  journal = {AI},
  year    = {2026},
  volume  = {7},
  number  = {3},
  pages   = {80},
  doi     = {10.3390/ai7030080}
}

@article{Su2025AutonomySecuritySurvey,
  author  = {Hang Su and Jun Luo and Chang Liu and Xiao Yang and Yichi Zhang and Yinpeng Dong and Jun Zhu},
  title   = {A Survey on Autonomy-Induced Security Risks in Large Model-Based Agents},
  year    = {2025},
  journal = {arXiv preprint arXiv:2506.23844},
  url     = {https://arxiv.org/abs/2506.23844}
}

@misc{OpenClawRepo2026,
  author       = {{OpenClaw}},
  title        = {{OpenClaw: Personal AI Assistant}},
  howpublished = {\url{https://openclaw.ai}},
  note         = {Accessed March 3, 2026},
  year         = {2026},
  url          = {https://openclaw.ai}
}

@article{Liu2026AgenticUrbanPlanning,
  author = {Liu, Rui and Zhe, Tao and Peng, Zhong-Ren and Catbas, Necati and Ye, Xinyue and Wang, Dongjie and Fu, Yanjie},
  title = {Urban Planning in the Age of Agentic AI: Emerging Paradigms and Prospects},
  journal = {SIGKDD Explorations Newsletter},
  year = {2026},
  volume = {27},
  number = {2},
  pages = {35--42},
  publisher = {Association for Computing Machinery},
  address = {New York, NY, USA},
  issn = {1931-0145},
  doi = {10.1145/3787470.3787474},
  url = {https://doi.org/10.1145/3787470.3787474},
  month = dec
}

@misc{liu2026cityeditinghierarchicalagentic,
      title={City Editing: Hierarchical Agentic Execution for Dependency-Aware Urban Geospatial Modification}, 
      author={Rui Liu and Steven Jige Quan and Zhong-Ren Peng and Zijun Yao and Han Wang and Zhengzhang Chen and Kunpeng Liu and Yanjie Fu and Dongjie Wang},
      year={2026},
      eprint={2602.19326},
      archivePrefix={arXiv},
      primaryClass={cs.MA},
      url={https://arxiv.org/abs/2602.19326}, 
}

@misc{zhe2025constraintawarerouterecommendationnatural,
      title={Constraint-Aware Route Recommendation from Natural Language via Hierarchical LLM Agents}, 
      author={Tao Zhe and Rui Liu and Fateme Memar and Xiao Luo and Wei Fan and Xinyue Ye and Zhongren Peng and Dongjie Wang},
      year={2025},
      eprint={2510.06078},
      archivePrefix={arXiv},
      primaryClass={cs.AI},
      url={https://arxiv.org/abs/2510.06078}, 
}

@misc{zhe2026robustefficienttoolorchestration,
      title={Robust and Efficient Tool Orchestration via Layered Execution Structures with Reflective Correction}, 
      author={Tao Zhe and Haoyu Wang and Bo Luo and Min Wu and Wei Fan and Xiao Luo and Zijun Yao and Haifeng Chen and Dongjie Wang},
      year={2026},
      eprint={2602.18968},
      archivePrefix={arXiv},
      primaryClass={cs.AI},
      url={https://arxiv.org/abs/2602.18968}, 
}

@inproceedings{yao2023react,
  title = {{ReAct}: Synergizing Reasoning and Acting in Language Models},
  author = {Yao, Shunyu and Zhao, Jeffrey and Yu, Dian and Du, Nan and Shafran, Izhak and Narasimhan, Karthik and Cao, Yuan},
  booktitle = {International Conference on Learning Representations (ICLR) },
  year = {2023},
  html = {https://arxiv.org/abs/2210.03629},
}

@inproceedings{schick2023toolformer,
  title={Toolformer: Language Models Can Teach Themselves to Use Tools},
  author={Schick, Timo and Dwivedi-Yu, Jane and Dess{\'i}, Roberto and Raileanu, Roberta and Lomeli, Maria and Hambro, Eric and Zettlemoyer, Luke and Cancedda, Nicola and Scialom, Thomas},
  booktitle={Advances in Neural Information Processing Systems (NeurIPS)},
  year={2023}
}

@article{wang2023voyager,
  title   = {Voyager: An Open-Ended Embodied Agent with Large Language Models},
  author  = {Guanzhi Wang and Yuqi Xie and Yunfan Jiang and Ajay Mandlekar and Chaowei Xiao and Yuke Zhu and Linxi Fan and Anima Anandkumar},
  year    = {2023},
  journal = {arXiv preprint arXiv: Arxiv-2305.16291}
}

@misc{nakano2022webgpt,
      title={WebGPT: Browser-assisted question-answering with human feedback}, 
      author={Reiichiro Nakano and Jacob Hilton and Suchir Balaji and Jeff Wu and Long Ouyang and Christina Kim and Christopher Hesse and Shantanu Jain and Vineet Kosaraju and William Saunders and Xu Jiang and Karl Cobbe and Tyna Eloundou and Gretchen Krueger and Kevin Button and Matthew Knight and Benjamin Chess and John Schulman},
      year={2022},
      eprint={2112.09332},
      archivePrefix={arXiv},
      primaryClass={cs.CL},
      url={https://arxiv.org/abs/2112.09332}, 
}

@misc{packer2024memgptllmsoperatingsystems,
      title={MemGPT: Towards LLMs as Operating Systems}, 
      author={Charles Packer and Sarah Wooders and Kevin Lin and Vivian Fang and Shishir G. Patil and Ion Stoica and Joseph E. Gonzalez},
      year={2024},
      eprint={2310.08560},
      archivePrefix={arXiv},
      primaryClass={cs.AI},
      url={https://arxiv.org/abs/2310.08560}, 
}

@misc{kaushik2014naturaluserinterfacestrend,
      title={Natural User Interfaces: Trend in Virtual Interaction}, 
      author={Dr. Manju Kaushik and Rashmi Jain},
      year={2014},
      eprint={1405.0101},
      archivePrefix={arXiv},
      primaryClass={cs.HC},
      url={https://arxiv.org/abs/1405.0101}, 
}

@ARTICLE{880078,
  author={Zue, V.W. and Glass, J.R.},
  journal={Proceedings of the IEEE}, 
  title={Conversational interfaces: advances and challenges}, 
  year={2000},
  volume={88},
  number={8},
  pages={1166-1180},
  doi={10.1109/5.880078}
}

@inproceedings{zou2019feedrec,
  author = {Zou, Lixin and Xia, Long and Ding, Zhuoye and Song, Jiaxing and Liu, Weidong and Yin, Dawei},
  title = {Reinforcement Learning to Optimize Long-term User Engagement in Recommender Systems},
  booktitle = {Proceedings of the ACM SIGKDD Conference on Knowledge Discovery and Data Mining (KDD)},
  year = {2019},
  pages = {2810--2818}
}

@inproceedings{zhao2018pagewise,
  author = {Zhao, Xiangyu and Xia, Long and Zhang, Liang and Ding, Zhuoye and Yin, Dawei and Tang, Jiliang},
  title = {Deep Reinforcement Learning for Page-Wise Recommendations},
  booktitle = {Proceedings of the ACM Conference on Recommender Systems (RecSys)},
  year = {2018},
  pages = {95--103}
}

@article{Lin_2024,
   title={A Survey on Reinforcement Learning for Recommender Systems},
   volume={35},
   ISSN={2162-2388},
   url={http://dx.doi.org/10.1109/TNNLS.2023.3280161},
   DOI={10.1109/tnnls.2023.3280161},
   number={10},
   journal={IEEE Transactions on Neural Networks and Learning Systems},
   publisher={Institute of Electrical and Electronics Engineers (IEEE)},
   author={Lin, Yuanguo and Liu, Yong and Lin, Fan and Zou, Lixin and Wu, Pengcheng and Zeng, Wenhua and Chen, Huanhuan and Miao, Chunyan},
   year={2024},
   month=oct, pages={13164–13184} }

@article{Huang_2025,
   title={A Survey on Hallucination in Large Language Models: Principles, Taxonomy, Challenges, and Open Questions},
   volume={43},
   ISSN={1558-2868},
   url={http://dx.doi.org/10.1145/3703155},
   DOI={10.1145/3703155},
   number={2},
   journal={ACM Transactions on Information Systems},
   publisher={Association for Computing Machinery (ACM)},
   author={Huang, Lei and Yu, Weijiang and Ma, Weitao and Zhong, Weihong and Feng, Zhangyin and Wang, Haotian and Chen, Qianglong and Peng, Weihua and Feng, Xiaocheng and Qin, Bing and Liu, Ting},
   year={2025},
   month=jan, pages={1–55} }

@article{10.1145/3571730,
author = {Ji, Ziwei and Lee, Nayeon and Frieske, Rita and Yu, Tiezheng and Su, Dan and Xu, Yan and Ishii, Etsuko and Bang, Ye Jin and Madotto, Andrea and Fung, Pascale},
title = {Survey of Hallucination in Natural Language Generation},
year = {2023},
issue_date = {December 2023},
publisher = {Association for Computing Machinery},
address = {New York, NY, USA},
volume = {55},
number = {12},
issn = {0360-0300},
url = {https://doi.org/10.1145/3571730},
doi = {10.1145/3571730},
journal = {ACM Comput. Surv.},
month = mar,
articleno = {248},
numpages = {38}
}

@inproceedings{10.1145/3711896.3736570,
author = {Mohammadi, Mahmoud and Li, Yipeng and Lo, Jane and Yip, Wendy},
title = {Evaluation and Benchmarking of LLM Agents: A Survey},
year = {2025},
isbn = {9798400714542},
publisher = {Association for Computing Machinery},
address = {New York, NY, USA},
url = {https://doi.org/10.1145/3711896.3736570},
doi = {10.1145/3711896.3736570},
booktitle = {Proceedings of the 31st ACM SIGKDD Conference on Knowledge Discovery and Data Mining V.2},
pages = {6129–6139},
numpages = {11},
location = {Toronto ON, Canada},
series = {KDD '25}
}

@misc{deng2023mind2webgeneralistagentweb,
      title={Mind2Web: Towards a Generalist Agent for the Web}, 
      author={Xiang Deng and Yu Gu and Boyuan Zheng and Shijie Chen and Samuel Stevens and Boshi Wang and Huan Sun and Yu Su},
      year={2023},
      eprint={2306.06070},
      archivePrefix={arXiv},
      primaryClass={cs.CL},
      url={https://arxiv.org/abs/2306.06070}, 
}

@inproceedings{zheng2024seeact,
  title={GPT-4V(ision) is a Generalist Web Agent, if Grounded},
  author={Boyuan Zheng and Boyu Gou and Jihyung Kil and Huan Sun and Yu Su},
  booktitle={Forty-first International Conference on Machine Learning},
  year={2024},
  url={https://openreview.net/forum?id=piecKJ2DlB},
}

@misc{zhou2024webarenarealisticwebenvironment,
      title={WebArena: A Realistic Web Environment for Building Autonomous Agents}, 
      author={Shuyan Zhou and Frank F. Xu and Hao Zhu and Xuhui Zhou and Robert Lo and Abishek Sridhar and Xianyi Cheng and Tianyue Ou and Yonatan Bisk and Daniel Fried and Uri Alon and Graham Neubig},
      year={2024},
      eprint={2307.13854},
      archivePrefix={arXiv},
      primaryClass={cs.AI},
      url={https://arxiv.org/abs/2307.13854}, 
}
